\documentclass{article}
\PassOptionsToPackage{numbers, compress}{natbib}

\usepackage[preprint]{neurips_2024}

\usepackage[utf8]{inputenc} 
\usepackage[T1]{fontenc}    
\usepackage{hyperref}       
\usepackage{url}            
\usepackage{booktabs}       
\usepackage{amsfonts}       
\usepackage{nicefrac}       
\usepackage{microtype}      
\usepackage[table]{xcolor}
\usepackage{xcolor}         
\usepackage{color}
\usepackage{graphicx}
\usepackage{amssymb}
\usepackage{wrapfig}
\usepackage{enumitem}
\usepackage{amsfonts}
\usepackage{amsmath}
\usepackage{algorithm}
\usepackage{algorithmic}
\usepackage{listings}
\usepackage[toc,title,page]{appendix}
\usepackage{graphicx}
\usepackage{multirow}
\usepackage{xspace}

\usepackage{pifont}
\newcommand{\True}{\textcolor{green}{\ding{51}}}
\newcommand{\False}{\textcolor{red}{\ding{55}}}

\usepackage{utfsym}
\definecolor{darkGreen}{RGB}{57,183,168}

\title{FullAnno: A Data Engine for Enhancing Image Comprehension of MLLMs}

\author{
Jing Hao$^1$\footnotemark[1]\thanks{Equal contribution.} \quad Yuxiang Zhao$^2$\footnotemark[1] \quad Song Chen$^2$\footnotemark[1] \quad Yanpeng Sun$^{2,3}$\footnotemark[1] \quad \textbf{Qiang Chen}$^2$ \quad \\ 
\textbf{Gang Zhang}$^2$ \quad \textbf{Kun Yao}$^2$ \quad \textbf{Errui Ding}$^2$ \quad \textbf{Jingdong Wang}$^2$ \\
\\
$^1$ The University of Hong Kong \\
$^2$ Baidu VIS \\
$^3$ Nanjing University of Science and Technology \\
\\
{\tt\small jinghao@connect.hku.hk} \\
{\tt\small \{zhaoyuxiang, chensong03, sunyanpeng, chenqiang13\}@baidu.com} \\
{\tt\small \{zhanggang03, yaokun01, dingerrui, wangjingdong\}@baidu.com}
}

\begin{document}
\maketitle
\begin{abstract}
 Multimodal Large Language Models (MLLMs) have shown promise in a broad range of vision-language tasks with their strong reasoning and generalization capabilities. However, they heavily depend on high-quality data in the Supervised Fine-Tuning (SFT) phase. The existing approaches aim to curate high-quality data via GPT-4V, but they are not scalable due to the commercial nature of GPT-4V and the simplicity of the prompts used to instruct the model. To this end, we devised the FullAnno system, which is a data engine that can generate large-scale, high-quality, and fine-grained image annotations consisting of the category and position of objects, region descriptions, text information, as well as image dense captions. This engine is characterized by its cascade annotation process, which involves multiple expert models and employs rich prompts to instruct LLMs in generating dense image captions. We re-annotated the COCO and Visual Genome datasets using our FullAnno system, tripling the number of object annotations and increasing the length of the original image captions by a factor of 15. Experiments show that the regenerated annotation can significantly enhance the capabilities of LLaVA-v1.5 on several benchmarks. The re-annotated data are available at \url{https://arcana-project-page.github.io}
\end{abstract}

\section{Introduction}
In the domain of large multi-modal models (LMMs), efficient modality alignment is critical but often constrained by the scarcity of high-quality image-text data \cite{llava, coco-llava, chen2023sharegpt4v, wang2023see, li2023silkie}. There is now a consensus that "quality over quantity" is particularly pertinent in training a versatile vision language model.
Experimental evidence has demonstrated that replacing the image-text pairs used in the SFT stage with equivalent comprehensive captions generated by the GPT-4 Vision model can lead to consistent performance gains across various LMMs and benchmarks \cite{chen2023sharegpt4v}.
However, the current mainstream image-text datasets often lack rich information and fine-grained semantics \cite{coco, VG}. These captions, usually brief and focus on prominent objects, result in a considerable reduction in information content and lead to sub-optimal modality alignment.
Currently, several initiatives \cite{chen2023sharegpt4v, chen2024allava} have been undertaken to generate high-quality image-text data, primarily relying on GPT-4 Vision model. However, these methods are not scalable due to the commercial nature of GPT-4 Vision and the simplicity of the prompts used to instruct the model. Consequently, the generated image captions depend entirely on the image understanding capabilities inhibited in the GPT-4 Vision model.

To address these issues and generate large-scale, high-quality, and fine-grained image caption datasets automatically, we designed the FullAnno data engine. This engine is characterized by its cascade annotation process involving multiple expert models and the use of rich prompts in instructing LLMs for generating image captions. These prompts include information about objects, positions, attributes, OCR, and region descriptions. 
The pipeline of our FullAnno data engine is demonstrated in Fig. \ref{fig:pipeline}. We employ the FullAnno engine to automatically re-annotate the COCO and Visual Genome datasets, tripling the number of object bounding boxes, providing OCR information, and increasing the token length of the original captions by 15 times.
To validate the effectiveness of re-annotated captions, we substitute our enhanced caption data with original annotations, without increasing the data volume, and keep the same model structure and training settings as LLaVA-v1.5-7B. A significant and consistent improvement is observed among many benchmarks, demonstrating the benefits of high-quality image captions for enhancing the capabilities of LMMs.

\section{Data Engine}

\begin{figure*}
	\centering
	\includegraphics[width=\linewidth]{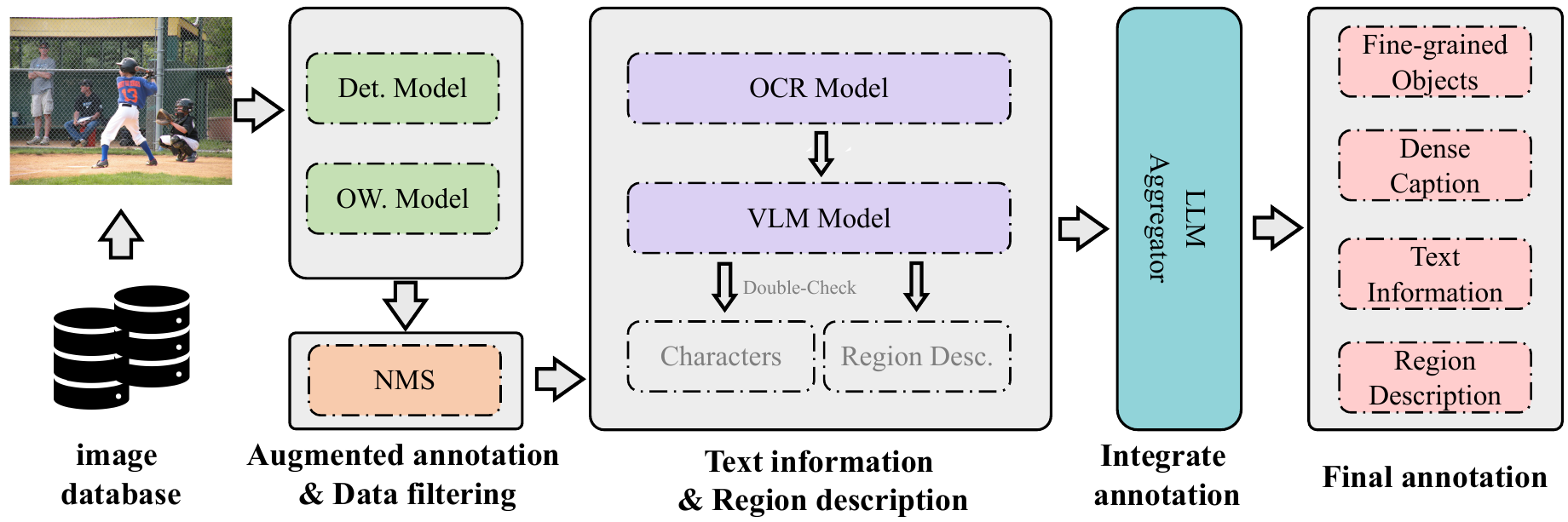}
	
	\caption{The \textbf{Arcana data engine} involves three crucial steps: \textbf{(1)} Augmenting and filtering image basic annotations, \textbf{(2)} obtaining text information and description for each annotated region, and \textbf{(3)} using a large language model to integrate these visual annotations into different types of captions.}
	\label{fig:pipeline}
\end{figure*}

The LLM model lacks the ability to comprehend image information, hence introducing an image translation task during the SFT stage accelerates MLLM's understanding of image content. The primary objective of the image translation task is to comprehensively describe all information within the image. However, existing caption data is overly simplistic, overlooking numerous important details within images, which hampers MLLM's comprehension of images. Therefore, we devised a data engine to acquire comprehensive annotations for images. The process, depicted in Fig.~\ref{fig:pipeline}, consists of three stages: \textbf{Augmented annotation and data filtering}, \textbf{text information and region description}, \textbf{integrate annotations} to obtain final annotation information.

\textbf{Augmented annotation and data filtering.} To augment initial annotation information, we utilize enhanced detection models~\cite{chen2023group,meng2021conditional}, and open-vocabulary detection models to extract text from images, pinpoint precise locations of objects, and identify all categories present within the images. Although augmented annotations obtained from specialized models offer comprehensive information, they are susceptible to noise and inaccuracies. To address this challenge, we implement a multifaceted filtering process to refine and eliminate unnecessary annotations. Specifically, we aggregate results using Non-Maximum Suppression (NMS) and apply thresholding to filter out noisy annotations. The IoU threshold for NMS is 0.75.


\textbf{Text information and Region description.} 
Text presented in images includes vital information for content analysis, which is also a basic element for image perception. 
We introduce an Optical Character Recognition (OCR) model ~\cite{sheng2021centripetaltext,zhu2021fourier} for obtaining text information contained in the image. To ensure the accuracy of the OCR information, we additionally employed the LLaVA-v1.5 to verify and correct the content of each detected OCR region.
At the same time, we established a matching relationship between OCRs and object regions by adhering to two criteria: the OCR bounding boxes should be completely contained within object regions, and subsequently choosing the smallest object region in terms of area to match each OCR entry.
Simultaneously, we generated region descriptions for each object using LLaVA-v1.5. To produce more accurate region descriptions, the visual prompt input to LLaVA-v1.5 consisted of the object region cropped from the whole image with a certain amount of surrounding context and the text prompt ``\textit{You glimpsed the image and saw a \{category\_name\}. Please describe the image in a few sentences:\ }''.

\begin{figure*}
	\centering
	\includegraphics[width=\linewidth]{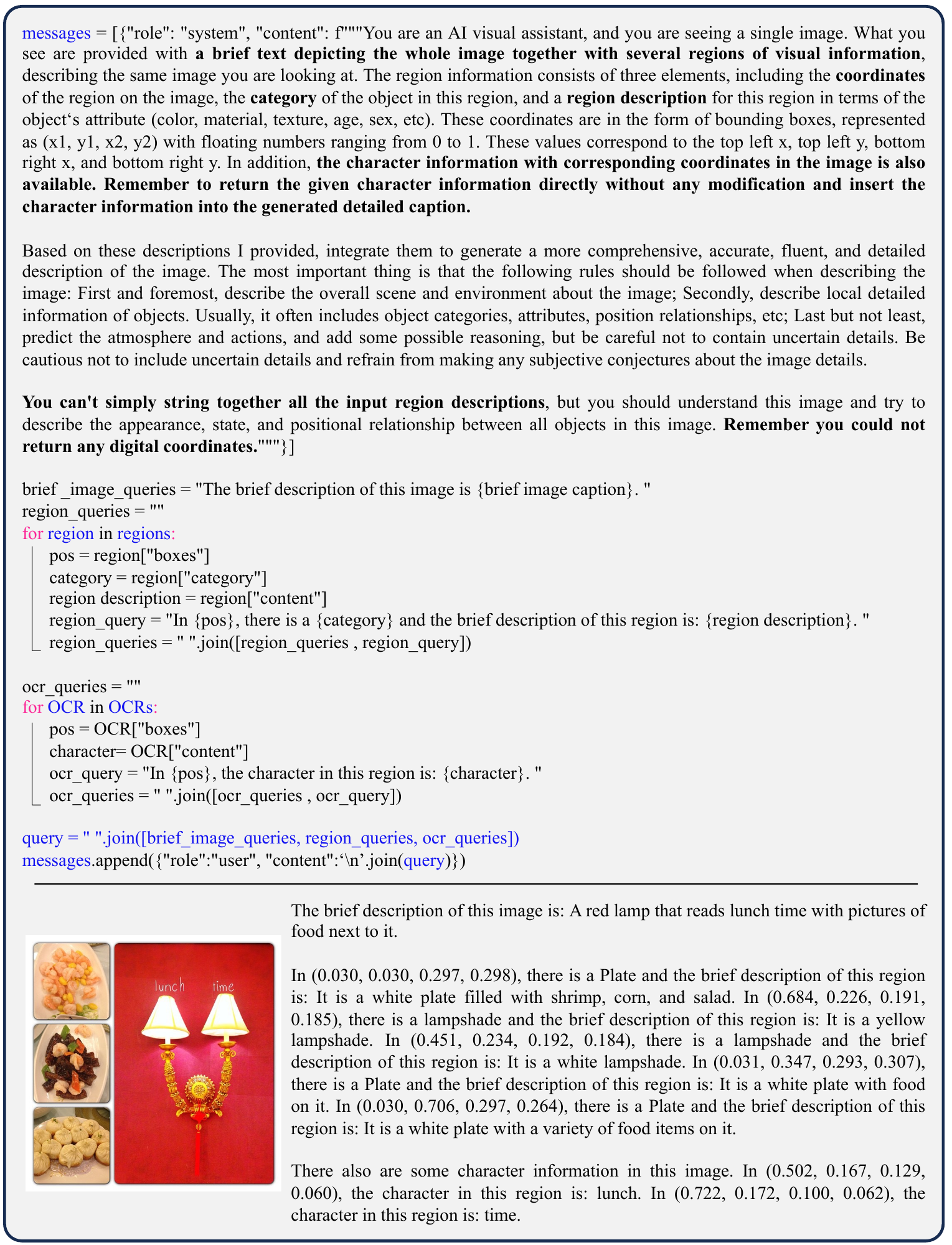}
	\caption{For each query, we illustrate the prompt construction process for instructing GPT-3.5. Note that ``message'' represents the final prompt. One example is displayed in the bottom row. The image is not visible to GPT-3.5 and is provided for reference only.}
	\label{fig:prompt}
\end{figure*}

\textbf{Integrating annotation.} To consolidate the discrete annotation results into a detailed caption for translating image content, we introduce a large language model, GPT-3.5. This model assists in integrating the aforementioned discrete annotation results to generate detailed captions for the images. Rather than instructing LLMs with simple prompts, our prompts include the category and position of objects, region descriptions, and text information within the image. The format of our prompt used to generate image dense caption from GPT-3.5 is shown in Fig. \ref{fig:prompt}.
These prior knowledge in the image provides richer and more detailed semantic information in textual form, enabling LLMs to fully understand the image and generate high-quality image captions. We also found that these prior knowledge mitigates the issue of model hallucinations to some extent, which will be discussed in Sec. \ref{3.2:dataset analysis}.

Ultimately, each image receives four types of visual annotations: dense caption, text information, object annotation, and region description. Here is a brief overview of each type:
\begin{itemize}
    \item Dense Caption: This detailed description of object attributes such as color, behavior, and relationships enhances the LMM's understanding of image content, potentially improving its ability to generate accurate and informative captions.
    \item Text Information: Incorporating text information present in images through OCR enhances the LMM's ability to describe textual elements such as signs, labels, or captions within the image, enriching the generated captions with textual context.
    \item Object Annotation: The positional information of objects obtained through detection enables the LMM to spatially ground its generated captions, ensuring that descriptions accurately correspond to the locations of objects in the image.
    \item Region Description: By providing information about specific regions in the image, this data component helps the LMM localize objects and understand their spatial relationships, contributing to the precision and coherence of generated captions.
\end{itemize}

\section{Dataset}
\subsection{Dataset Overview}

\begin{table}[h]
\centering
\setlength\tabcolsep{4pt}
\renewcommand\arraystretch{1.3}
\caption{Annotation Comparison of the COCO and Visual Genome datasets. The ``Cap'' refers to the ``caption'', and the ``ATL'' abbreviates the ``Average Token Length''. The token length is counted by the tokenlizer of LLaMa \cite{touvron2023llama}.}
\label{tab:Comparison_cocoVisual Genome}
\resizebox{1\textwidth}{!}{%
\begin{tabular}{l|cccc|cccccc}
\hline
Dataset & Simple Cap  & Dense Cap & Region Cap & OCR  & \# Images & \# Boxes & ATL for Dense Cap & ATL for Region Cap \\
\hline
COCO \cite{coco} & \checkmark & \usym{2613}  & \usym{2613} & \usym{2613}   & 118k   & 0.86M & 11.94 & -   \\
Visual Genome \cite{VG} & \checkmark & \usym{2613}  & \usym{2613} & \checkmark  & 76k & 0.61M & - & 2.5 \\ 
LLaVA-ReCap \cite{coco-llava} & \checkmark & \checkmark & \usym{2613} & \usym{2613} & 118k   & 0.86M & 196.12 & - \\
Ours  & \checkmark & \checkmark & \checkmark & \checkmark & 180k & 4.16M & 181.07 & 42.79 \\

\hline
\end{tabular}
}
\end{table}

We update annotations on COCO and Visual Genome datasets in terms of bounding box, as well as region description, and we also supplement extra fundamental elements in images like dense captions and OCR information that do not exist in the official annotations. 
First and foremost, we replenish more objects and increase the number of annotated bounding boxes of objects threefold, from 1.47M to 4.16M. 

Simultaneously, we generated region descriptions for each object using LLaVA-v1.5. To produce more accurate region descriptions, the visual prompt input to LLaVA-v1.5 consisted of the object region cropped from the whole image with a certain amount of surrounding context, and the text prompt is: ``\textit{You glimpsed the image and saw a \{category\_name\}. Please describe the image in a few sentences:\ }''. The final average length of tokens for the region captions is 42.79.
Besides, we augmented the OCR information in the COCO and Visual Genome datasets, with a total of 58k entries. At the same time, we established a matching relationship between OCRs and object regions by adhering to two criteria: the OCR bounding boxes should be completely contained within object regions, and subsequently choosing the smallest object region in terms of area to match each OCR entry.

Utimately, for dense caption generation, we utilized LLM to integrate prior information, including categories and positions of objects, sampled captions, region descriptions, and OCR, to generate comprehensive dense captions. By explicitly providing contextual priors from the image, LLM can focus more on the integration task, which is relatively simpler compared to image translation tasks, and also mitigate the issue of model hallucinations to some extent.
The comparison of our generated data with COCO, Visual Genome, and LLaVA-ReCap-COCO118k is presented in Table \ref{tab:Comparison_cocoVisual Genome}.

\begin{figure*}
	\centering
	\includegraphics[width=\linewidth]{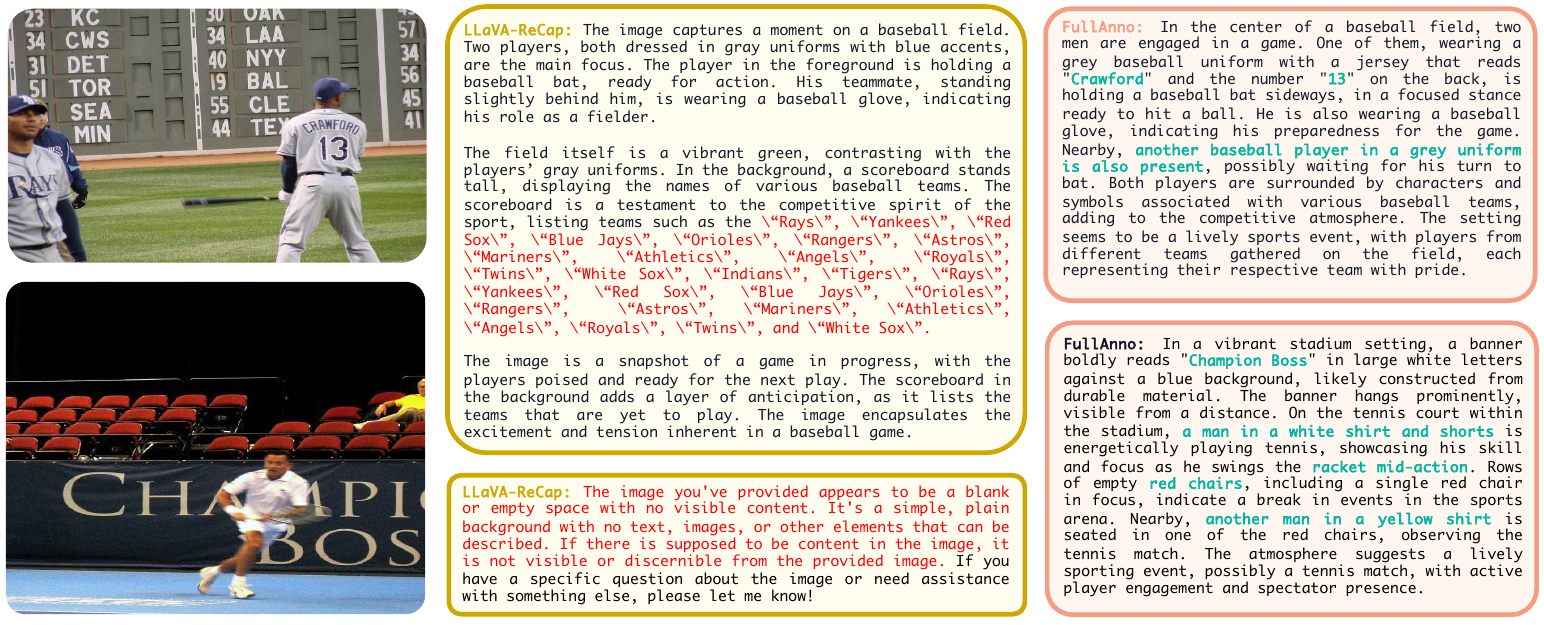}
	\caption{Comparisons of dense caption between LLaVA-RaCap-COCO118k \cite{coco-llava} and ours. The hallucination parts are highlighted in \textcolor{red}{red}, whereas detailed and accurate parts are emphasized in \textcolor{darkGreen}{dark green}. }
	\label{fig:llava_recap}
\end{figure*}

\subsection{Dataset Analysis}
\label{3.2:dataset analysis}
Figure \ref{fig:llava_recap} presents the comparison of dense captions between ours and those generated by LLaVA-ReCap using LLaVA-NeXT-34B. One obvious shortcoming of LLaVA-ReCap is the hallucination issue when abbreviations appear in the image, whereas this issue does not occur in the results generated by our FullAnno engine. Besides, our results include textual information from the images, such as "13" and "Carwford," which is not present in LLaVA-ReCap.
Simultaneously, LLaVA-ReCap sometimes includes some failed cases, such as the bottom row in Figure \ref{fig:llava_recap}. However, our FullAnno engine consistently outputs information about objects, attributes, colors, and OCR in the image, thanks to its ability to generate dense captions through a cascade approach. 

\begin{figure}[h]
\centering
\includegraphics[width=\linewidth]{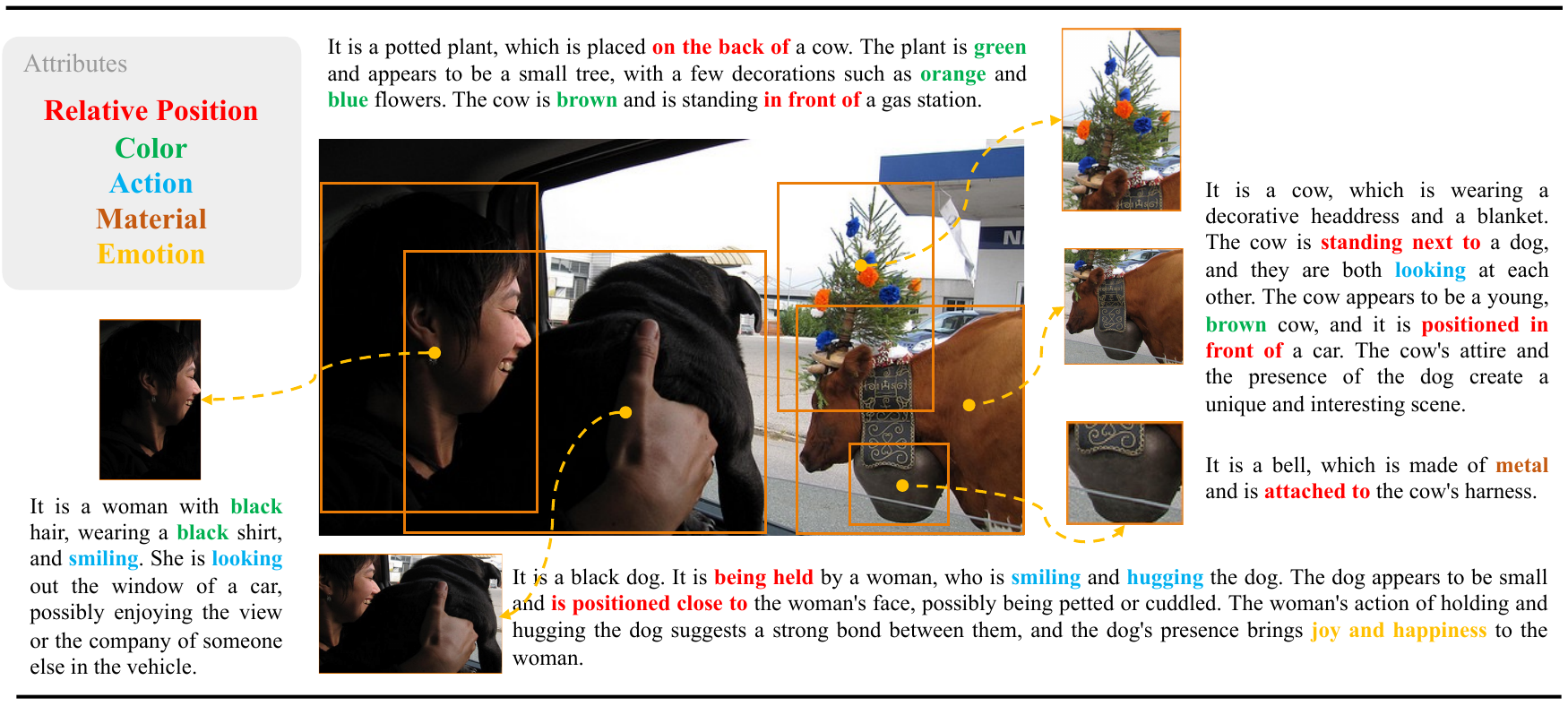}
 \vspace{-4mm}
\caption{The region description includes various object attributes such as relative position, color, action, material, and emotion. Best viewed in color.}
\label{fig:region_attributes}
\end{figure}

We also found that the region description generated for each object includes various object attributes such as relative position, color, action, material, and emotion, which are shown in Fig. \ref{fig:region_attributes}. These detailed pieces of information can provide more fine-grained prior knowledge for generating image dense captions, thereby ensuring the correctness and granularity of the dense captions.

\subsection{Effectiveness of Dataset}

\begin{figure}[h]
\centering
\includegraphics[width=\linewidth]{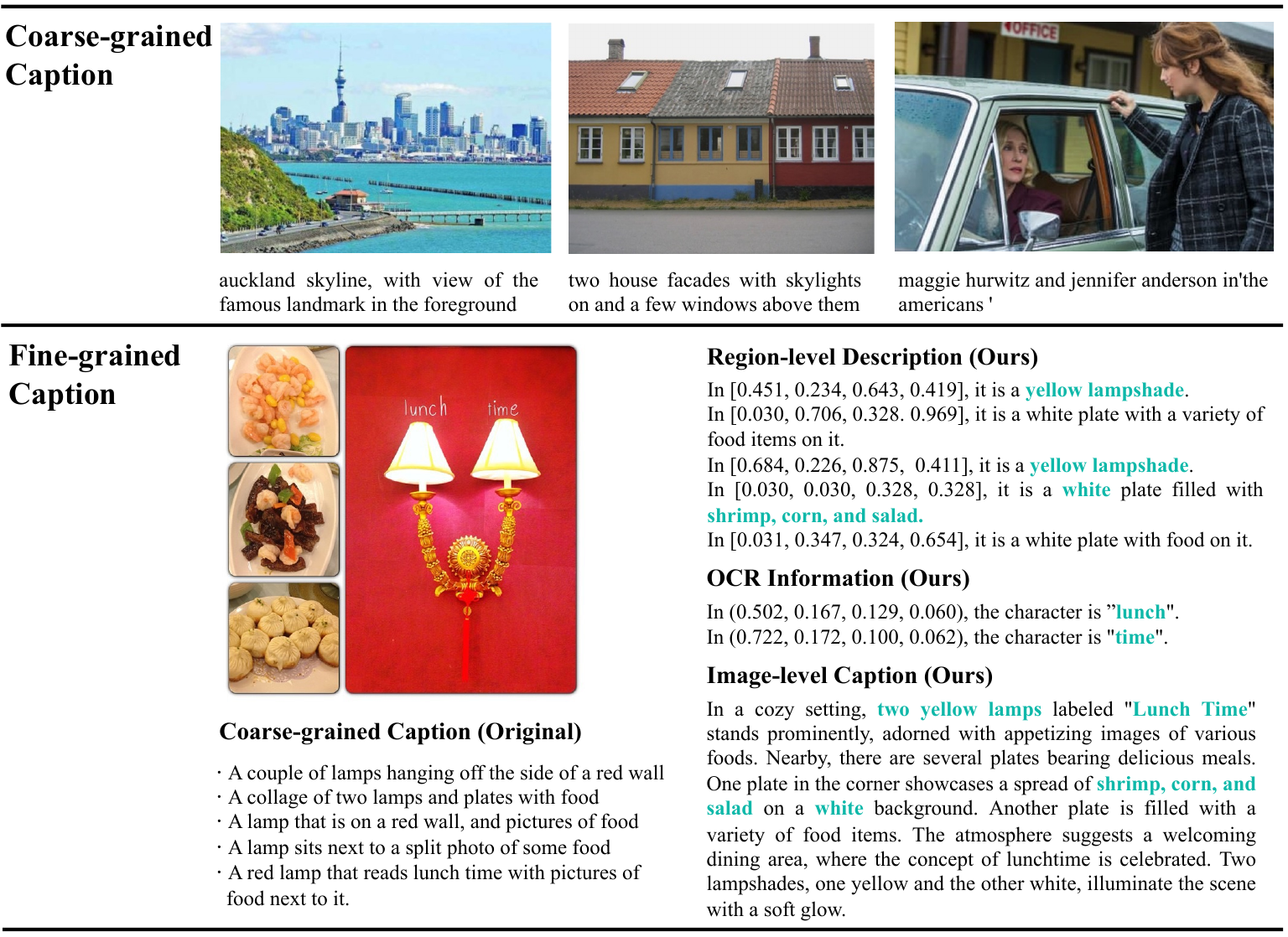}
 \vspace{-4mm}
\caption{The coarse-grained caption v.s. Our fine-grained caption used in pre-training stage of LLaVA. Important visual recognition-related description are highlighted in \textcolor{darkGreen}{dark green}.}
\label{fig:caption_comparision}
\end{figure}

In Fig.~\ref{fig:caption_comparision}, we compare our enhanced caption data with the original caption data used by LLaVA-v1.5-7B~\cite{liu2023llava1.5} in the pre-training stage.
It is clearly that our enhanced annotations provide a more fine-grained description of the images, thereby improving the model's visual perception without increasing the data volume.
To validate the effectiveness of enhanced annotations, we ensured the use of the same model structure and training method as LLaVA-v1.5-7B~\cite{liu2023llava1.5}, and introduced our enhanced caption data. The results are presented in the Table.~\ref{tab:train_data}.
We observe a significant improvement simply by incorporating our enhanced caption data into the pre-training stage, demonstrating the benefits of fine-grained image descriptions for enhancing model visual perception.

\begin{table}[h]
\centering
\setlength\tabcolsep{4pt}
\renewcommand\arraystretch{1.3}
\caption{Influence of the incorporation of the detailed caption data.}
\label{tab:train_data}
\resizebox{0.8\textwidth}{!}{%
\begin{tabular}{c|cccccc}
\hline
  \textbf{Detailed Caption}& SQA\_I  & TextVQA & POPE & MME  & MM-Vet & SEED  \\
\hline
\False & 66.8 & 58.2  & 85.9 & 1510.7   & 31.1   & 58.6   \\
\True  & \textbf{69.6}\textcolor{red}{$_{\uparrow2.8}$} & \textbf{59.4}\textcolor{red}{$_{\uparrow1.2}$}  & \textbf{86.6}\textcolor{red}{$_{\uparrow0.7}$} & \textbf{1519.1}\textcolor{red}{$_{\uparrow8.4}$}  & \textbf{31.4}\textcolor{red}{$_{\uparrow0.3}$}  & \textbf{62.1}\textcolor{red}{$_{\uparrow4.5}$} \\ \hline
\end{tabular}%
}
\end{table}

\section{Conclusion}
We designed a FullAnno system, which is a data engine that can generate large-scale, high-quality, and fine-grained image caption datasets automatically. 
Besides, FullAnno can also provide a diverse range of information present in images, including the category and position of objects, region descriptions, and OCR information.
These prior knowledge in images also used to instruct LLMs for generation detailed image caption and could mitigate the issue of hallucinations to some extent. Experiments on LLaVA-v1.5 demonstrate the significant and consistent improvement among many benchmarks, proving the effectiveness of our  FullAnno data engine. Hoping for our FullAnno system could further propel the advancement of LMMs from the hith-quality data generation perspective. dgenerate large-scale, high-quality, and fine-grained image caption datasets automatically generate large-scale, high-quality, and fine-grained image caption.

{\small
\bibliographystyle{ieee_fullname}
\bibliography{egbib.bib}
}
\end{document}